# Early- and in-season crop type mapping without current-year ground truth: generating labels from historical information via a topology-based approach

Chenxi Lin[a], Liheng Zhong[b], Xiao-Peng Song[c], Jinwei Dong[d], David B. Lobell[e], Zhenong Jin[a,*]

[a] Department of Bioproducts and Biosystems Engineering, University of Minnesota Twin Cities, St Paul, MN, United States

[b] Ant Group, Beijing, China

[c] Department of Geosciences, Texas Tech University, Lubbock, TX, United States

[d] Key Laboratory of Land Surface Pattern and Simulation, Institute of Geographic Sciences and Natural Resources Research, Chinese Academy of Sciences, Beijing, China

[e] Department of Earth System Science, Center on Food Security and the Environment, Stanford University, Stanford, CA, United States

corresponding authors:  *jinzn@umn.edu (Jin Z.)

**Abstract**

Land cover classification in remote sensing is often faced with the challenge of limited ground truth labels. Incorporating historical ground information has the potential to significantly lower the expensive cost associated with collecting ground truth and, more importantly, enable early- and in-season mapping that is helpful to many pre-harvest decisions. In this study, we propose a new approach that can effectively transfer knowledge about the topology (i.e. relative position) of different crop types in the spectral feature space (e.g. the histogram of SWIR1 vs RDEG1 bands) to generate labels, thereby support crop classification in a different year. Importantly, our approach does not attempt to transfer classification decision boundaries that are susceptible to inter-annual variations of weather and management, but relies on the more robust and shift-invariant topology information. We tested this approach for mapping corn/soybeans in the US Midwest and paddy rice/corn/soybeans in Northeast China using Landsat-8 and Sentinel-2 data. Results show that our approach automatically generates high-quality labels for crops in the target year immediately after each image becomes available. Based on these generated labels from our approach, the subsequent crop type mapping using a random forest classifier can reach the F1 score as high as 0.887 for corn as early as the silking stage and 0.851 for soybean as early as the flowering stage and the overall accuracy of 0.873 in the test state of Iowa. In Northeast China, F1 scores of paddy rice, corn and soybeans and the overall accuracy can exceed 0.85 two and half months ahead of harvest. Overall, these results highlight unique advantages of our approach in transferring historical knowledge and maximizing the timeliness of crop maps. Our approach supports a general paradigm shift towards learning transferrable and generalizable knowledge to facilitate land cover classification.

# 1. Introduction

Crop mapping, i.e. identifying what crops are growing in farmers' fields, is fundamental for a wide range of applications in agriculture, such as supply chain management, development intervention, crop insurance and risk management (Benami et al., 2020; Lobell et al., 2015). With the capacity to repeatedly and consistently monitor the terrestrial surface, remote sensing has become an effective tool for cropland mapping over the past three decades (Weiss et al., 2020). In recent years, the application of remote sensing has been further promoted due to rapid growth in data volume and accessibility, as well as development in sophisticated algorithms to extract useful information. The bottleneck for producing high-resolution, high-accuracy classification maps thus has shifted from a lack of data or image processing capacity towards a lack of ground truth (Rußwurm et al., 2020).

Ground truth, also known as labels, provides reliable references for classifier training and validation in remote sensing-based supervised classifications. Lack of ground truth can lead to poorly performed classifiers. While being important for supervised classifications, collecting appropriate ground truth is a major challenge. Ideally, ground truth should be obtained from surveys that collect targeted, first-hand information (Jin et al., 2019; Song et al., 2017; Xiong et al., 2017; Zhang et al., 2019), but these efforts are often hindered by the substantial cost of time and labor, particularly in underdeveloped countries. A compromise is to use well-recognized official products, such as the United States Department of Agriculture (USDA)'s Cropland Data Layer (CDL) (Boryan et al., 2011) as the weak "ground truth" (Cai et al., 2018; Johnson and Mueller, 2021; Wang et al., 2019; Xu et al., 2021, 2020). Furthermore, this type of periodically updated maps are only available in very few countries, and the mapping procedure may take months or even years to finish. For instance, the CDL map is generated based on satellite imagery over an entire growing season and annually updated ground truth, and thus is not released until January/February of the following year (Boryan et al., 2011). In summary, lack of timely ground truth is a major hurdle to the early- and in-season crop mapping (Cai et al., 2018; Johnson and Mueller, 2021; Skakun et al., 2017).

To map crop types in the target year that has little or even no ground truth, several methods have been proposed to transfer historical knowledge by first training a classifier using historical labels and then applying it to the target year (Cai et al., 2018; Ghazaryan et al., 2018; Hao et al., 2020, 2016; Johnson and Mueller, 2021; Konduri et al., 2020; Waldner et al., 2015; Wang et al., 2019; Xu et al., 2021, 2020; Yaramasu et al., 2020; You and Dong, 2020; Zhong et al., 2014). These methods are known as "decision boundary-based approaches" because they attempted to transfer decision boundaries built from historical years to the target year. Fig. 1 summarizes generalized procedures of these decision boundary-based approaches using corn and soybean classification as an example. The first step is to select representative training samples based on certain spectral features (here the red-edge (RDEG1) band and shortwave infrared (SWIR1) band from Sentinel-2 as an example) (Figs. 1a, b). Next, samples are input to a classifier to build decision boundaries. Finally, decision boundaries built from historical data are applied to new data in the target year. In practice, a more complex feature space is usually employed, including multiple bands and other derived metrics, and the classification algorithms can be hard margin-based such as support vector machine (SVM) or soft probability-based such as logistic regression and neural networks (Liu et al., 2011).

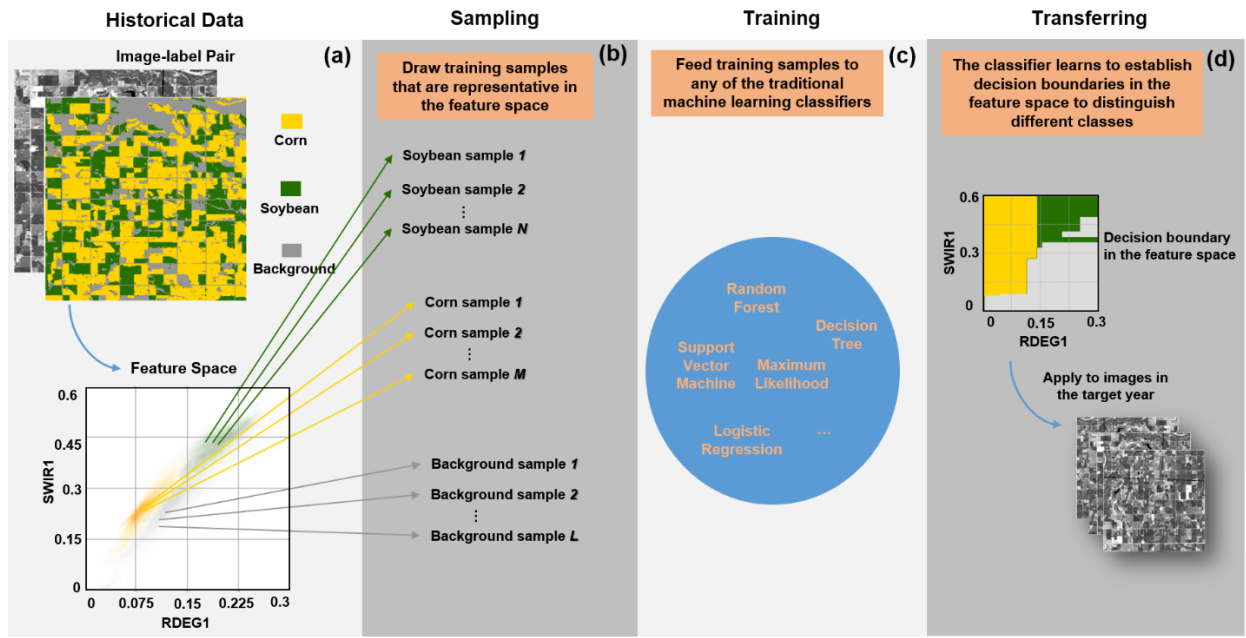

**Fig. 1.** The workflow of decision boundary-based approaches. (a) and (b) Select representative training samples for corn, soybean and background in the RDEG1/SWIR1 feature space from historical satellite imagery and labels. (c) and (d) Train classifiers using selected training samples to build decision boundaries in the feature space and transfer it to the target year to classify corn and soybeans.

The underlying assumption for transferring decision boundaries is stable and consistent spectral features over time. However, this assumption is often violated, particularly for land cover types with rapid seasonal changes under cross-year scenarios because spectral features have both intra-annual variability and inter-annual variability due to changes in weather and crop progress (Zhong et al., 2014). Therefore, decision boundaries are often not directly applicable when transferring over time, especially to another year. This is true even with incorporating phenological features that are believed to reduce the impacts of intra- or inter-annual variability (Hao et al., 2020, 2015; Liu et al., 2019; Zhong et al., 2012, 2011), thus existing methods often don't perform well in early- and in-season mappings (Johnson and Mueller, 2021). Fig. 2a provides an example showing the intra- and inter-annual variability (2018 and 2019) in the Sentinel-2 RDEG1/SWIR1 feature space of corn and soybean fields from a subregion in Iowa. In contrast, Fig. 2b highlights an interesting and time-invariant pattern such that the soybean cluster is constantly on the northeast of the corn cluster starting from early July regardless of their absolute positions and shapes in the RDEG1/SWIR1 feature space in both 2018 and 2019. Our preliminary examination of multi-year data indicates that for each year, there will be a certain period when a stable topological relationship can be easily identified by human eyes regardless of inter-annual variability (Fig. S1). Theoretically, such a stable topological relationship between corn and soybeans can be transferred from historical years to the target year, and likely for other crops as well.

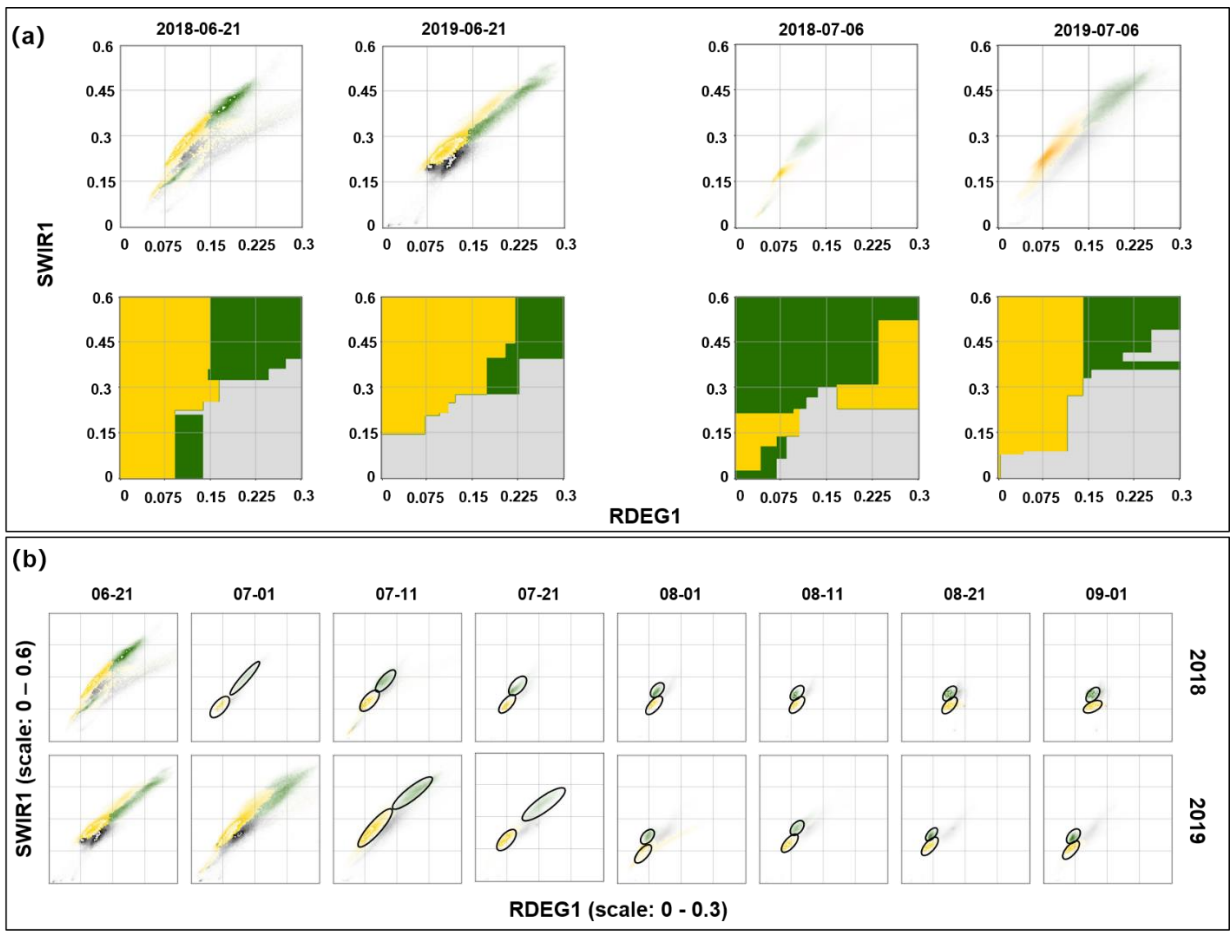
5

**Fig. 2.** Intra- and inter-annual variability in the Sentinel-2 RDEG1/SWIR1 feature space from a subregion in Iowa (top left: 43.225N/96.639W, bottom right:43.087N/96.502W). Feature values of corn (yellow) and soybean (green) in the same region change over time and hard decision boundaries which are built from feature values are different at different dates. (a) Comparison of feature values and decision boundaries at Jun 21st and Jul 6th, 2018 and 2019. (b) Change of topology relationship between corn (yellow) and soybeans (green) during the growing season in 2018 and 2019. Black denotes the background.

This study proposes an approach to effectively transfer knowledge about the topology (e.g. relative position) of different crop types in the spectral feature space so as to generate high-quality labels in the target year. Once training labels are available, there is a wide range of algorithm choices for the subsequent classification task, and here we choose random forest (RF) for its simplicity. We tested this framework for mapping corn/soybeans in the US Midwest and paddy rice/corn/soybeans in Northeast (NE) China. Experiments covered different sensors, years, landscapes and crop diversities and strongly supported our ultimate goal to generalize this approach to many other crop types as well as to other regions where collecting ground truth remains a major challenge. It should be noted that the novelty of this study is a topology-based approach to generate labels that combined with a simple RF classifier already showed a good performance in early season crop type mapping, thus laying a foundation for others to integrate with more advanced classification algorithms and develop an end-to-end framework.

## 2. Method

### 2.1. Study area and data

#### 2.1.1. Study area

In this study, we selected the state of Iowa in the US Midwest and Songnen Plain in NE China as two examples to test the proposed topology-based approach (Fig. 3). Iowa is a representative agricultural state in the US Midwest, which had 36% of land covered by corn and 25% covered by soybeans (NASS, 2019) and the production of corn and soybeans ranked first and second in the US in 2020, respectively (USDA, 2020). Although post-season classifications can often achieve high accuracy in this corn/soybeans dominant region, early- and in-season prediction of crop types remain a challenging task (Johnson and Mueller, 2021). Songnen Plain is an important food production region of China and its dominant crops are corn, soybeans and paddy rice (Yang et al., 2017; Zhang et al., 2018). Existing crop type maps covering this region were generated based on field-collected ground truth from 2017 - 2019 and a post-season classification method (You et al., 2021). The two study areas have different climate conditions, landscapes, as well as crop diversities and therefore, are selected to validate the generalizability of the proposed approach.

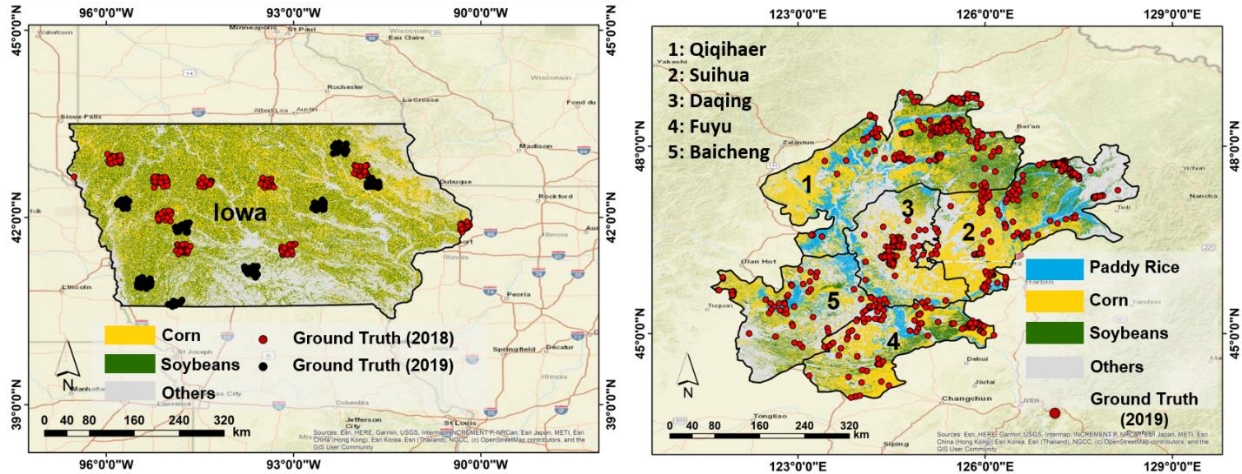

**Fig. 3.** Locations of study areas and field-collected ground truth samples (left: Iowa in the US, right: Songnen Plain in NE China). The crop type information is derived from the 2019 CDL in Iowa and the 2019 cropland map in NE China.

**2.1.2. Satellite data**

To test the utility of the proposed approach with different sensors, we collected Sentinel-2 Multispectral Instrument (MSI) L1C Top of Atmosphere (TOA) reflectance (COPERNICUS/S2) and Landsat-8 Operational Land Imager (OLI) surface reflectance (SR) data (LANDSAT/LC08/C01/T1_SR) through Google Earth Engine (GEE) (Gorelick et al., 2017). The Sentinel-2 L1C product has been processed with radiometric and geometric corrections and cloud-contaminated pixels were removed. We used L1C instead of more well-calibrated L2A because the L2A product is only available for 2019 in the study area. Similarly, Landsat-8 data in GEE is well-processed with atmospheric correction and includes a quality band derived from the CFMask algorithm to filter out cloud and cloud shadow etc. (Foga et al., 2017; Zhu and Woodcock, 2012). For both sensors, we collected images from June 1st (DOY 151) to approximately Sep 30th (DOY 270) for multiple years. For Sentinel-2 and Landsat-8, we generated 5-day and 16-day composite as a "time step", respectively. Finally, we ended up with 24 time steps for Sentinel-2 for 2017, 2018 and 2019, respectively, and 8 time steps for Landsat-8 from 2013 to 2019, respectively.

**2.1.3. Ground truth data**

We used the pixel-wise CDL in Iowa (Boryan et al., 2011) and cropland map in NE China (You et al., 2021) during the training stage and field-collected ground truth for validation. We acknowledge that the CDL and cropland map in NE China are not perfect ground truth as they are predictions to approximate real land covers. However, because the proposed approach requires a large number of samples to learn the topology knowledge between crop types, using these pixel-wise products to help acquire sufficient training samples. This also demonstrates the high tolerance of the proposed approach to classification errors in the training stage. we collected independent ground truth samples for Iowa in 2018 and 2019 and for NE China from 2017 to 2019 to assess the mapping accuracy (Table 1). We followed the approach designed in Song et al., (2017) and collected crop type information over random samples over the Conterminous United States in 2018 and 2019. The subsets of the national samples located in the State of Iowa were used in this study. In NE China, samples covering three target crop types (i.e., paddy rice, corn and soybeans) as well as other land cover types (e.g., grassland, wetland, build-up etc.) were recorded with a mobile GIS device and had a double check against satellite imagery after the fieldwork (You et al., 2021). To distinguish these two data sources, we hereafter refer to the predicted products, i.e., the CDL and cropland map in NE China, as the pseudo ground truth and the in situ data as the field-collected ground truth.

**Table 1**
The number of ground truth samples in Iowa (2018&2019) and NE China (2019).

| Crop Type | Iowa | | NE China |
|---|---|---|---|
| | 2018 | 2019 | 2019 |
| Corn | 67 | 55 | 129 |
| Soybean | 55 | 55 | 163 |
| Paddy rice | \ | \ | 169 |
| Others | 49 | 40 | 212 |

**2.2. Experiment overview**

Our approach aims to generate high-quality labels for different crop types in the target year by effectively transferring topology knowledge between crops (here corn, soybeans and paddy rice). We conducted four experiments to evaluate the effectiveness of the approach regarding different years, sensors, regions and crop types (Table 2). Note that all experiments were performed at the 30m resolution level and thus all Sentinel-2 imagery was resampled to 30m.

**Table 2**
Summary of designed experiments

| Experiment | Crop Types | Time Period | Satellite Data | Region |
|---|---|---|---|---|
| 1 | Corn Soybeans Others | Training: 2017&2018 Testing: 2019 | Sentinel-2 MSI L1C | Iowa |
| 2 | Corn Soybeans Others | Training: 2017&2019 Testing: 2018 | Sentinel-2 MSI L1C | Iowa |
| 3 | Corn Soybeans Others | Training: 2013-2017 Testing: 2018&2019 | Landsat-8 OLI SR | Iowa |
| 4 | Paddy Rice Corn Soybeans Others | Training: 2017&2018 Testing: 2019 | Sentinel-2 MSI L1C | NE China |

Figs. 4a, b use corn/soybeans mapping in Iowa with Sentinel-2 imagery as an example to illustrate the workflow. In the training stage, we use cropped image patches and their corresponding pseudo ground truth patches of all time steps in historical years (2018 in this example) to generate 2-dimensional (2D) histograms (Fig. 4a). The 2D histograms are hereafter called heat maps and details regarding generating heat maps from satellite image patches are given in section 2.3. Next, we train a convolutional neural network (CNN) model to recognize topology relationships between corn and soybean in the heat map. In the application stage, the model trained with historical data is applied to each time step in the target year (2019 in this example) to generate high-quality labels for corn and soybean (Fig. 4b, section 2.4). High-quality labels generated from the proposed approach are fed into a classification algorithm to map crop types in the target year (Fig. 4c, section 2.5). We use a RF classifier because the focus of this paper is the proposed topology-based method that can generate labels in the target year, but does not include testing the performance of advanced classification algorithms in early- and in-season crop type mapping.

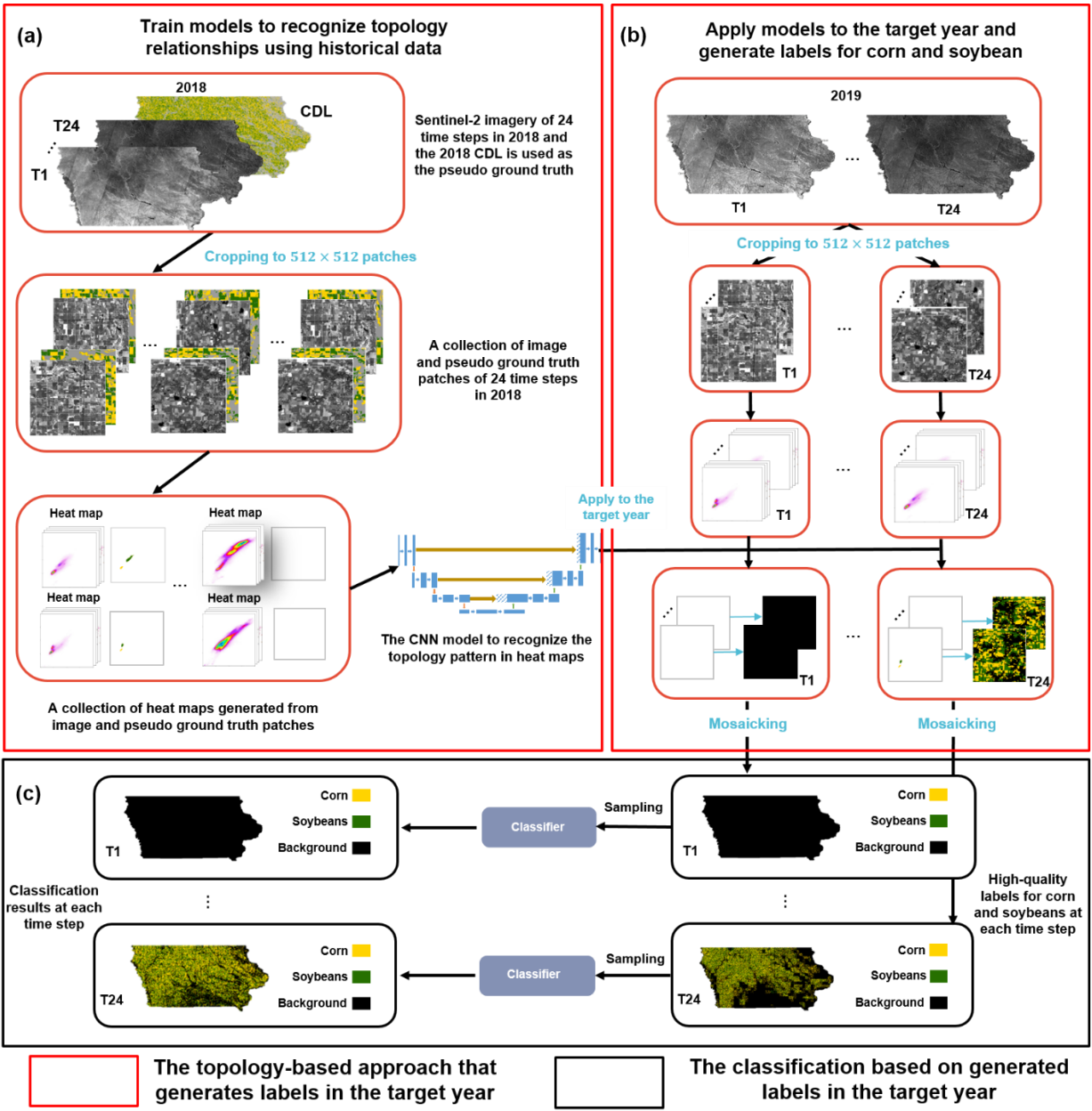

**Fig. 4.** Flowchart of the proposed topology-based approach to generate labels ((a) and (b)) and support crop type mapping in the target year ((c)). (a) A convolutional neural network(CNN) model is trained with historical data to recognize topology relationships in heat maps. (b) The CNN model is applied to each time step in the target year to generate high-quality labels for corn and soybeans. (c) The generated labels assist further crop type mapping in the target year.

### 2.3. Generating heat maps

The heat map in this study specifically refers to the location and density of pixels in any 2D feature space, e.g., the feature space of RDEG1 and SWIR1 in Fig. 5. To obtain a heat map, a satellite image is divided into 512×512 patches and all pixels are projected to the 2D feature space according to the value of the two features (Fig. 5a, b). In this step, three parameters are specified: bins, scales for the x-axis and y-axis. The example in Fig. 5 uses a bin size of 128 and a scale of 0.3/0.6 for the x/y axis. The rule for selecting bins and scales is to ensure the heat map with high quality and appropriate image size. For example, 1) a bin size of 64 is not appropriate because it will make the heat map too coarse; 2) a bin size of 256 will unnecessarily increase the computation load because 128 is fine enough to show sufficient information; 3) A scale of 1/1 is not a good choice because the clusters in Fig. 5b will be shifted to the bottom left corner, thus impairing the quality of the heat map. In general, visualizations and more detailed discussion about the selection of bin sizes and scales are given in Fig. S2.

Heat maps generated from satellite images were divided into two categories based on whether or not they present the desired topology relationship, i.e., type-I denotes heat maps containing a recognizable topology relationship and type-II represents those that do not (Figs. 5c, d). To speed up the categorization, we first used the JM distance (Dong et al., 2013; Murakami et al., 2001) to pre-categorize all heat maps into the two types, and then conducted a visual inspection to correct any errors in the pre-categorized heat maps. More details about the JM distance can be found in section S1 in the supplementary material. Note that though the JM distance was employed to facilitate the categorization process, it is not necessarily required. Relying only on visual interpretation on all heat maps (though can be time-consuming) or using other measurements, e.g., the Mahalanobis distance (Mahalanobis, 1936) and Kullback-Leibler divergence (Kullback and Leibler, 1951), can achieve the same goal in this process.

The process of defining targets (corn and soybeans pixel to be predicted in the heat map) for heat maps is explained in Figs. 5c, d. Here, we used the 50% target, i.e., the top 50% of corn and soybean distributions in the heat map, because the purpose of this step was to extract very "pure" and typical crop pixels for later classification. To illustrate this, we tested both 50% and 100% targets and found that the 50% target eliminates the overlap between corn and soybeans that occurs in the 100% target (the black region). A 3D visualization showing the difference between the 100% and 50% target is presented in Fig. S3. Targets for type-II heat maps are blank images, indicating that the model is not expected to learn any knowledge from these inputs.

To facilitate the model's learning of the topological relationship, we added three other channels (similar concept to bands of satellite imagery) in addition to the heat map (Fig. 5e). The second channel shows the location and density of pixels in the satellite image patch that have never been cultivated with target crops, e.g., corn and soybean in this example, according to historical information. The third channel records the X coordinates of each nonzero pixel in the 1$^{st}$ channel and therefore, pixels have incrementally increasing values horizontally but keep constant vertically. The forth channel records the y coordinates of each nonzero pixel in the 1$^{st}$ channel and has constant values along the x-axis. By stacking four channels in the input image, we expect the model to mainly learn the topological relationship from the 1$^{st}$ channel whereas the other three channels boost the model performance by confining the potential trajectory of corn and soybean clusters.

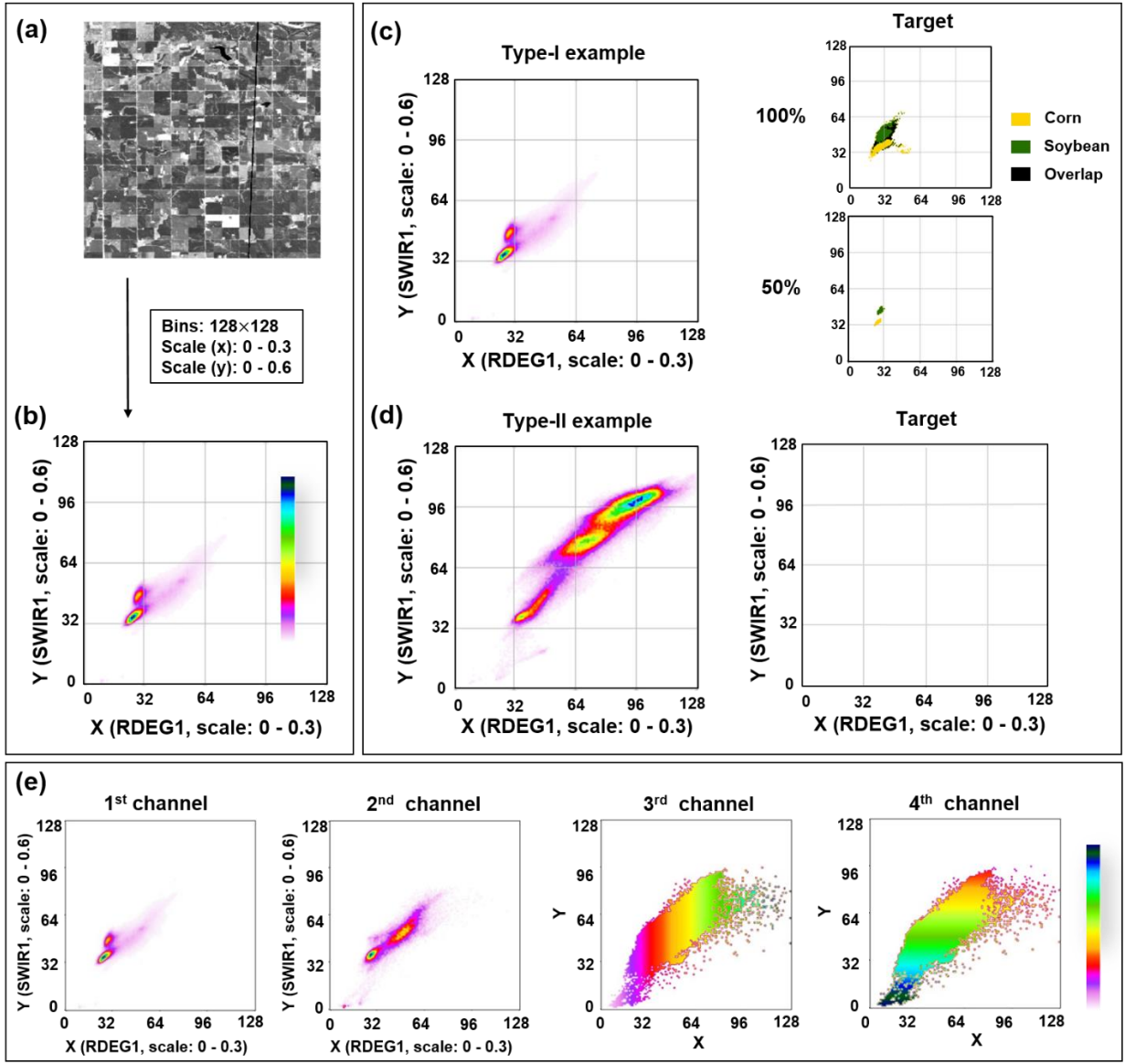

**Fig. 5.** Generating heat maps and targets. (a) The satellite image patch with the size of 512 × 512. (b) The heat map with a size of 128 × 128 showing the location and density of pixels in the satellite imagery in the 2D feature space. (c) For heat maps with recognizable topology patterns (type-I), pixels contributing to the most confident 50% of the distribution of each class in the heat map will be labelled; (d) For heat maps without recognizable topology patterns (type-II), nothing will be labelled and blank images will be used instead; (e) Four channels of the input. The 1st channel is the heat map of the satellite image patch; the 2nd channel shows the location and density of pixels in the satellite image patch that have never been cultivated with target crops; the 3rd and 4th channels indicate the X and Y coordinates of all nonzero pixels in the 1st channel.

## 2.4. Generating high-quality labels in the target year

With the four-channel input image and targets generated in section 2.3, the task becomes a supervised classification where the model is trained to learn the location of crops in the 2D feature space. Therefore, we chose a popular CNN model, i.e., U-Net (Ronneberger et al., 2015), to perform this task (Fig. 6a). The architecture of the U-Net is provided in Fig.S3. At the application stage, the U-Net model was applied to the target year to transfer the historical topology knowledge and recognize crops (Fig. 6b). Since the recognized crops were in the 2D feature space, it required converting them to their spatial distribution in the satellite image patch, which was an inverse process of Fig. 5a, b. After this conversion, we could obtain label patches or blank patches according to whether or not crops have been recognized in the input images (Fig. 6c). Label patches provided high-quality labels for the crop classification in the target year (section 2.5). In the upper panel of Fig. 6c, not all corn and soybeans were recognized because we only used the 50% target in the training stage, but this missing is not vital as long as the recognized labels were sufficient to support further classifications. Note that labels were accumulated when imagery from later time steps became available and started to generate more labels. The methodology for accumulating labels are described in Fig. S4.

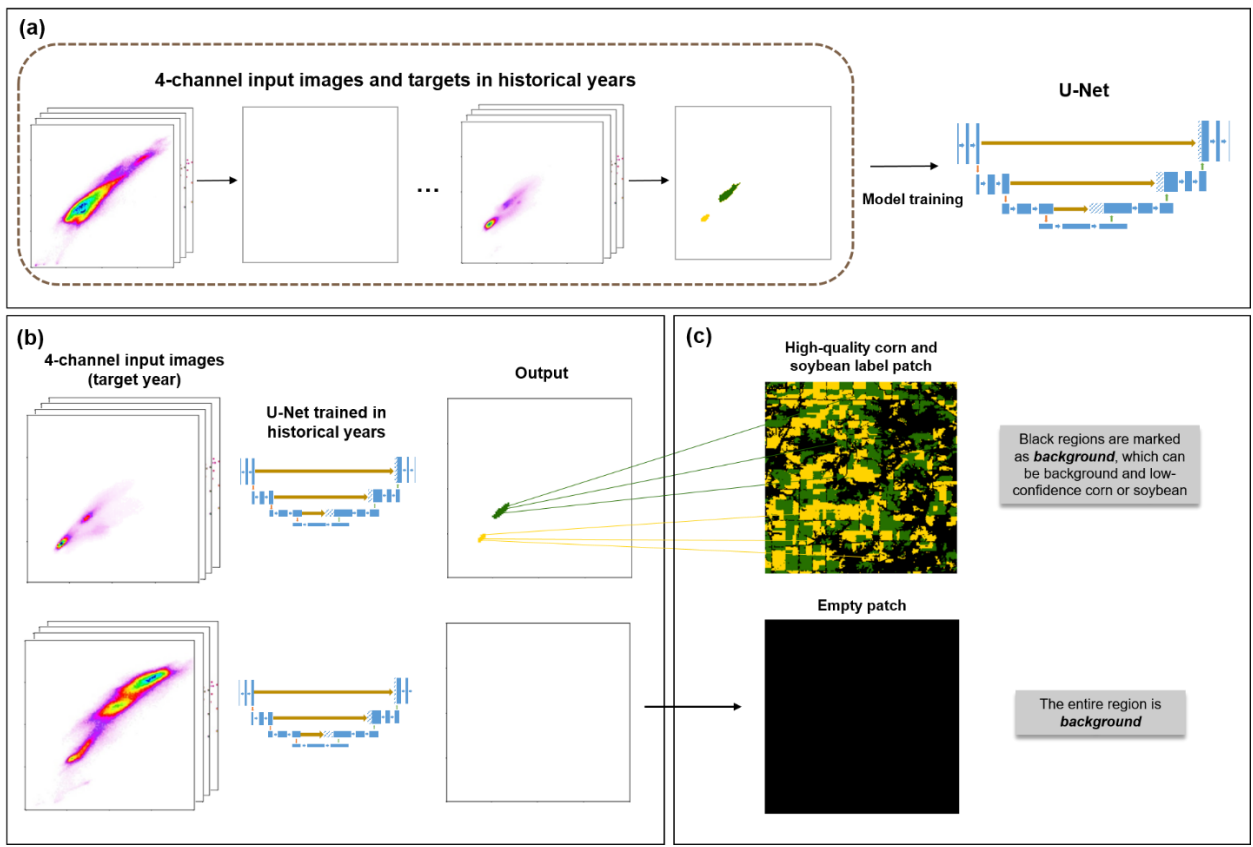

**Fig. 6.** The procedure of applying the U-Net model trained with historical topology knowledge to the target year to generate labels. (a) The 4-channel input and target in historical years was input to the U-Net to train it to recognize crops in the 2D feature space; (b) The 4-channel input generated in the target year was input to the U-Net model trained with historical knowledge; (c) Output images either contained recognized corn and soybean clusters or were blank depending on whether or not the input heat map showed desired topological relationships. Recognized corn and soybeans on the output image were converted to their spatial distributions on the satellite image.

## 2.5. Crop type classification in the target year

In addition to high-quality labels for target crops generated from the proposed topology-based approach, to perform crop type classification in the target year, background labels were also collected from historical information. For example, to map corn and soybeans in Iowa, background labels were selected from regions that had never been cultivated with corn and soybeans based on historical CDL. A RF classifier was selected and the input features include all spectral bands plus four derived metrics, i.e., the normalized difference vegetation index (NDVI) (Tucker, 1979), the enhanced vegetation index (EVI) (Huete et al., 2002), the green chlorophyll vegetation index (GCVI) (Gitelson et al., 2003) and the land surface water index (LSWI) (Xiao et al., 2002) to map crop types across the entire study region (here Iowa and NE China). Note that more advanced classification algorithms can be incorporated into the proposed framework but that is beyond the scope of this study as was discussed in section 4.1. For each time step with available crop labels, image composites created from all images within each 20-day period beginning from DOY 91 (April 1st) to the current date were used for classification. The 20-day period for image composite was selected after testing different period options (i.e., 5 days, 10 days, 15 days and 20 days). A shorter composite period (e.g., 5 and 10 days) was more susceptible to cloud contamination, thus leading to poorer classification performance. The 15-day period and 20-day period had comparable results but the 20-day image composite was more time-efficient as fewer features were used. More details about image compositing are presented in Fig. S5.

As the classification was performed at large scales, spatial variability was one major influential factor to the performance. To mitigate the impact of spatial variability, we split each study region into 9 grids since smaller subregions generally had smaller spatial variability than a larger region. For each grid, we randomly selected 2,000 labels for each crop from all labels generated from our proposed approach to train the RF classifier. Parameters for the classifier can be found in Table S1. Labels in each grid were prioritized for that grid and labels in the nearest neighbor grid were used if no labels are in the current grid (Fig. S6). Eventually, one classification map was generated for each time step if labels are available at that time step.

The accuracy for generated labels was evaluated by the pseudo ground truth, i.e., the CDL in Iowa and the cropland map in NE China of the target year and classification results were assessed by the field-collected ground truth. We used pseudo ground truth for evaluating generated labels because the field-collected ground truth, which was sampled randomly, largely did not overlap with the labels derived from the topology approach that aimed to generate the most representative labels. Although we acknowledge the inherent errors in the pseudo ground truth that were likely propagating to the evaluation, comparing them could also convey information regarding the approximate accuracy of generated labels. The number of generated labels and their agreement with pseudo ground truth were reported and F1 score and overall accuracy (OA) were used to assess the accuracy of crop type mapping using generated labels. The definition of the F1 score is:

$$F1 = \frac{2 \times UA \times PA}{UA + PA} \quad (1)$$

where UA and PA denote the user accuracy and producer accuracy.

## 3. Result

### 3.1. Mapping corn and soybean in Iowa

#### 3.1.1. Employed feature spaces and topology relationships in corn/soybeans mapping

In this experiment, we trained models to generate labels for corn and soybeans from the RDEG1/SWIR1 and NIR/SWIR1 feature space of Sentinel-2 and the NIR/SWIR1 feature space of Landsat-8. In the REDG1/SWIR1 feature space, we observed that corn pixels were first at the bottom left of soybean pixels and then moved directly below soybeans during the growing season (Fig. 7a). Here the topology relationship is not confined to one fixed type of relative position but rather any trajectory of relative positions that can be used to separate crop clusters. For example, corn pixels moved from the bottom right of soybean pixels, to directly below soybeans and finally to the bottom left of them in the NIR/SWIR1 feature space of Sentinel-2 (Fig. 7b). The NIR/SWIR1 feature space presented the same topology relationship between corn and soybeans as that of Sentinel-2 (Fig. 7c).

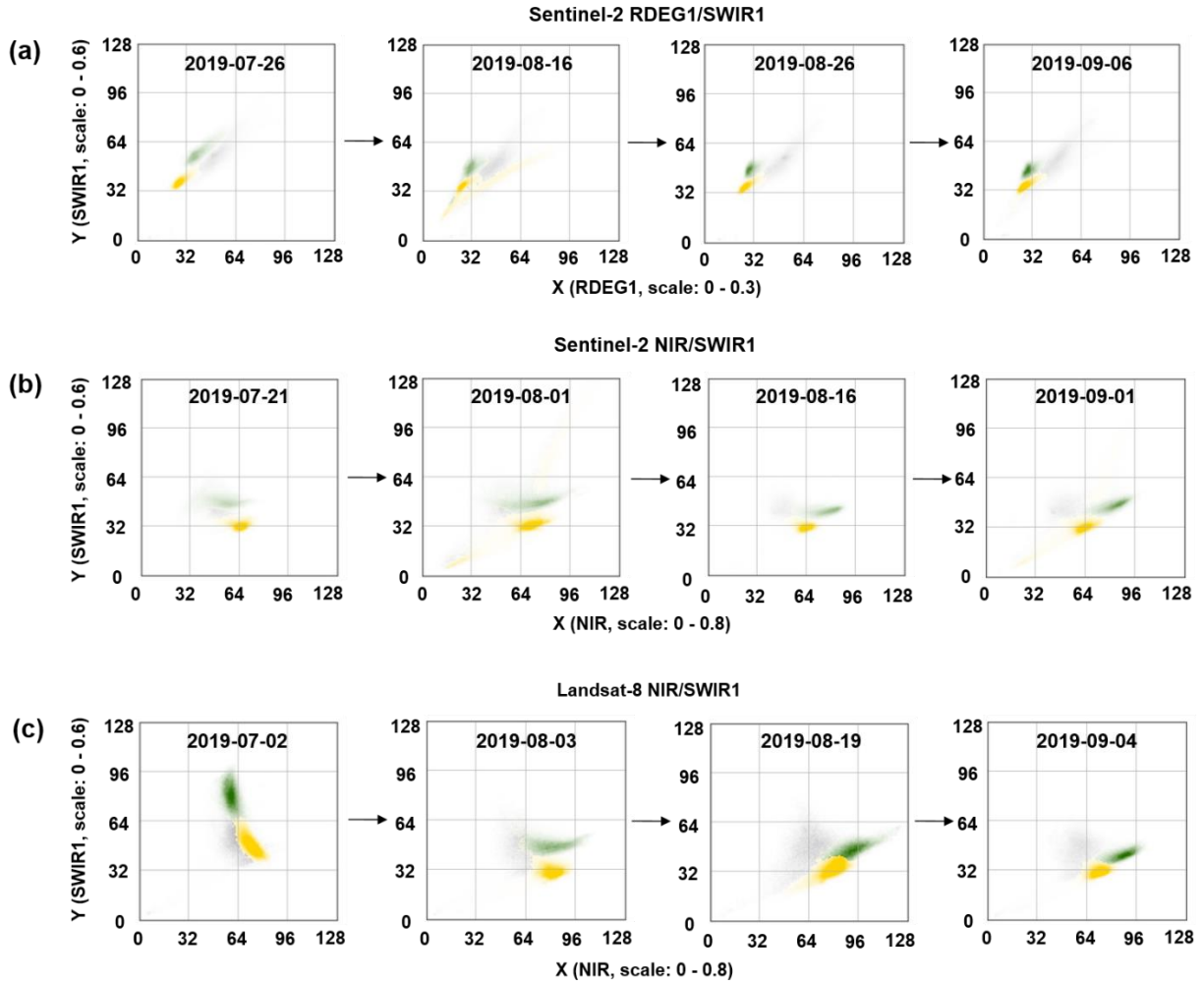

**Fig. 7.** Employed topology relationships between corn (yellow) and soybeans (green). Arrows indicate the progression of the growing season. Note that the date showing the topology relationship can have yearly variation depending on weather conditions, management practice, etc. (a) When using the RDEG1/SWIR1 feature space of Sentinel-2, the corn was first at the bottom left of soybeans and then moved directly below soybeans. (b) When using the NIR/SWIR1 feature space of Sentinel-2, the corn was first at the bottom right of soybeans, then moved directly below soybeans and finally moved to the bottom left of them. (c). When using the NIR/SWIR1 feature space of Landsat-8, it has the same topology relationship as that of Sentinel-2.

3**.1.2. Agreements and numbers of generated corn and soybean labels**

In the RDEG1/SWIR1 feature space of Sentinel-2 (Figs. 8a-d), the approach started to generate labels for both corn and soybeans at DOY 166 (Jun 15th) in 2018 and quickly achieved high agreements evaluated by the 2018 CDL. Average agreements for corn and soybean labels were 0.959 and 0.939, respectively. The number of both corn and soybean labels had sharp increases between DOY 181 (Jun 30th) and DOY 186 (Jul 5th) and slowed down until reaching a plateau. The number of labels accumulated to 27,664,780 and 25,716,243 for corn and soybean by the end of September. In the abnormal year of 2019, average agreements for corn and soybean labels remained as high as 0.990 and 0.960, respectively; and the number of labels reached 10,197,157 and 11,692,947 for the two crops, respectively. The first date starting to generate labels in 2019 was later than that in 2018 because of the delayed crop planting and growth (NASS, 2019).

The topology relationships in the NIR/SWIR1 feature space are more complex, e.g., corn and soybeans have more variable shapes and are less concentrated, compared to the RDEG1/SWIR1 feature space (Figs. 7a, b). This caused average agreements for corn and soybean labels to be lower than those derived from the REDG1/SWIR1 feature space, i.e., 0.873 and 0.868 in 2018 and 0.908 and 0.890 in 2019 for corn and soybean labels, respectively. The final numbers of labels were 38,992,283 in 2018 and 31,496,613 in 2019 for corn and were 28,951,304 in 2018 and 18,739,388 in 2019 for soybeans. The numbers of generated corn and soybean labels from the NIR/SWIR1 feature space were greater than those from the RDEG1/SWIR1 feature space, likely because of the higher prediction errors.

Experiments with Landsat-8 also showed promising results in extracting high-quality corn and soybean labels (Figs. 8i-l) and had comparable average agreements for the two crops (i.e., 0.937 and 0.847 in 2018 and 0.913 and 0.859 in 2019). However, influenced by the relatively sparse temporal frequency of Landsat-8, fewer labels were accumulated over time compared to Sentinel-2. Consequently, final label numbers for the two crops were 18,873,488 and 6,396,045 in 2018 and 26,180,567 and 11,889,137 in 2019.

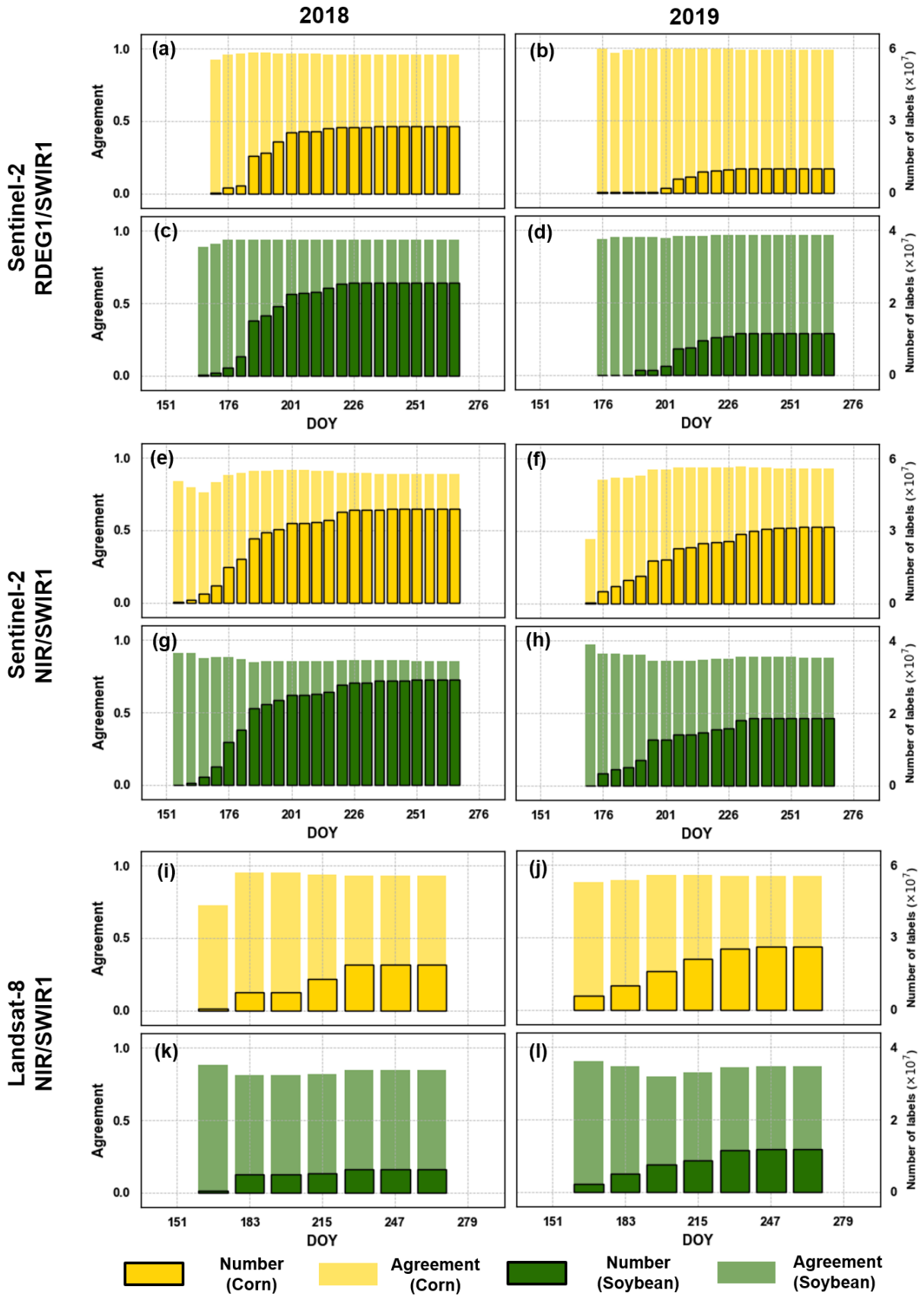

**Fig. 8.** Agreements and numbers of corn and soybean labels in Iowa, the US. (a)-(b) Corn labels in 2018 and 2019 generated from the RDEG1/SWIR1 feature space of Sentinel-2. (c)-(d) Soybean labels in 2018 and 2019 generated from the RDEG1/SWIR1 feature space of Sentinel-2. (e)-(f) Corn labels in 2018 and 2019 generated from the NIR/SWIR1 feature space of Sentinel-2. (g)-(h) Soybean labels in 2018 and 2019 generated from the NIR/SWIR1 feature space of Sentinel-2. (i)-(j) Corn labels in 2018 and 2019 generated from the NIR/SWIR1 feature space of Landsat-8. (k)-(l) Soybean labels in 2018 and 2019 generated from the NIR/SWIR1 feature space of Landsat-8.

### 3.1.3. Accuracy of corn and soybean classification

For simplification, we use the "RDEG1/SWIR1 experiment" and the "NIR/SWIR1 experiment" to name classifications using labels generated from the RDEG1/SWIR1 and NIR/SWIR1 feature space.

Using the field-collected ground truth as reference, both RDEG1/SWIR1 and NIR/SWIR1 experiments using Sentinel-2 had improved performance as the growing season progressed (Fig. 9). Both corn F1, soybean F1 and OA exceeded/approximated 0.85 by DOY 191 (Jul 10th) in 2018 in the RDEG1/SWIR1 experiment. The three metrics kept increasing before DOY 216 (Aug 4th) and stabilized at the maximum levels of 0.932, 0.855 and 0.860, respectively. The NIR/SWIR1 experiment in 2018 had similar maximum accuracy of 0.932, 0.864 and 0.865 for corn F1, soybean F1 and OA, respectively. The performance of the two experiments were comparable to the F1 scores and OA of the 2018 CDL (0.899, 0.855 and 0.854 for corn F1, soybean F1 and OA, respectively). In 2019, all metrics were lower in the early season and did not surpass 0.85 until DOY 206 (Jul 25th). The maximum F1 scores of corn and soybean and OA were 0.885, 0.860 and 0.873 in the RDEG1/SWIR1 experiment and 0.917, 0.923 and 0.907 in the NIR/SWIR1 experiment. Compared to the 2019 CDL (0.929, 0.901 and 0.90 for corn/soybean F1 scores and OA), the RDEG1/SWIR1 experiment was 3.9%, 4.1% and 2.7% lower in the three evaluation metrics, whereas the NIR/SWIR1 experiment had comparable results and was even 2.2% and 0.7% higher in the soybean F1 score and OA.

The classification accuracy for Landsat-8 NIR/SWIR1 experiment was slightly worse than that of the Sentinel-2 NIR/SWIR1 experiment (Fig. 9g-i). The maximum OA was 0.847 and 0.895 in 2018 and 2019, respectively. The poorer performance of Landsat-8 compared to Sentinel-2 could be attributed to the 16-day revisit frequency. One consequence of this sparser temporal resolution was fewer cumulative labels, which reduced the spatial diversity of corn and soybeans and caused the classifier's capacity to interpret spatial variability weakened. In addition, fewer images were available over the growing season for classification and therefore, the classifier was more vulnerable to outlier values, such as cloud-contaminated pixels.

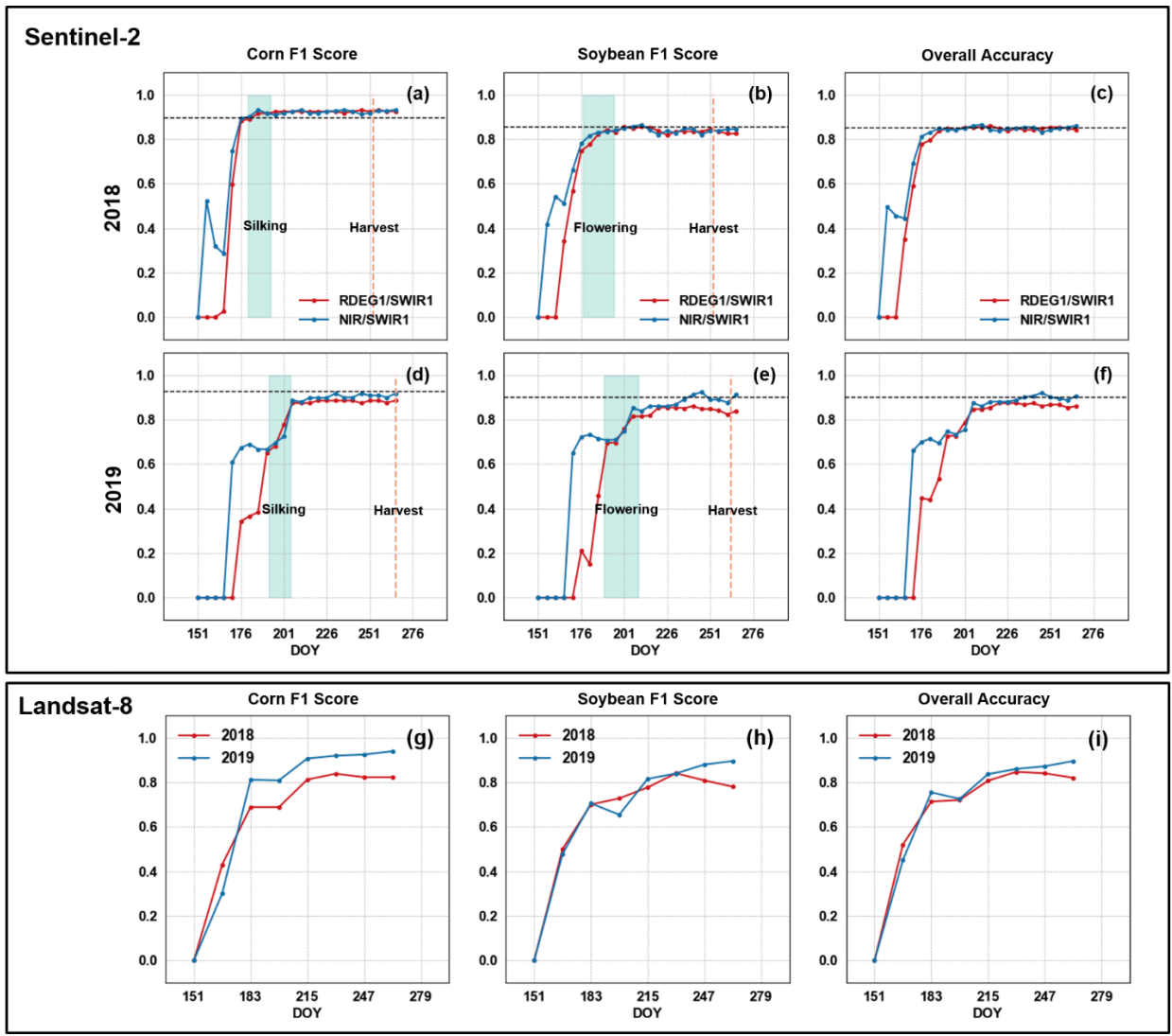

**Fig. 9.** Date-to-date comparison between corn F1 score, soybean F1 score and overall accuracy of classification results. Shaded regions denote dates when 25% and 75% percent of corn in the silking stage or soybean in the flowering stage, orange dash lines indicate NASS reported start date of harvest and black dash lines indicate the accuracy of the CDL. Zero values indicate that no labels were generated at the date. (a) F1 score of corn in 2018 using Sentinel-2. (b) F1 score of soybeans in 2018 using Sentinel-2. (c) Overall accuracy in 2018 using Sentinel-2. (d) F1 score of corn in 2019 using Sentinel-2. (e) F1 score of soybeans in 2019 using Sentinel-2. (f) Overall accuracy in 2019 using Sentinel-2. (g) F1 score of corn in 2018/2019 using Landsat-8. (h) F1 score of soybeans in 2018/2019 using Landsat-8. (i) Overall accuracy in 2018/2019 using Landsat-8.

### 3.2. Mapping paddy rice, corn and soybean in NE China

#### 3.2.1. Employed feature spaces and topology relationships in paddy rice/corn/soybeans mapping

As the objective of this experiment is to test the generalizability of the approach to map other crop types in a different landscape, we did not test the topology relationship in multiple feature spaces but only used the RDEG1/SWIR1 feature space of Sentinel-2. Two topology relationships were used to generate labels for the three crops, i.e., the relationship between corn and soybeans and between paddy rice and corn. Since we have already generated sufficient labels for the three crops using the two topology relationships, the relationship between paddy rice and corn was no longer needed. The same topology relationship between corn and soybeans as described in section 3.1.1 was used in this experiment (Fig. 10a). The topology relationship between paddy rice and corn presented in Fig.10b shows that paddy rice pixels can always be observed below corn pixels during the growing season no matter how they change positions. Consequently, we trained two U-Net models in this experiment to recognize topology relationships for paddy rice/corn and corn/soybeans, respectively.

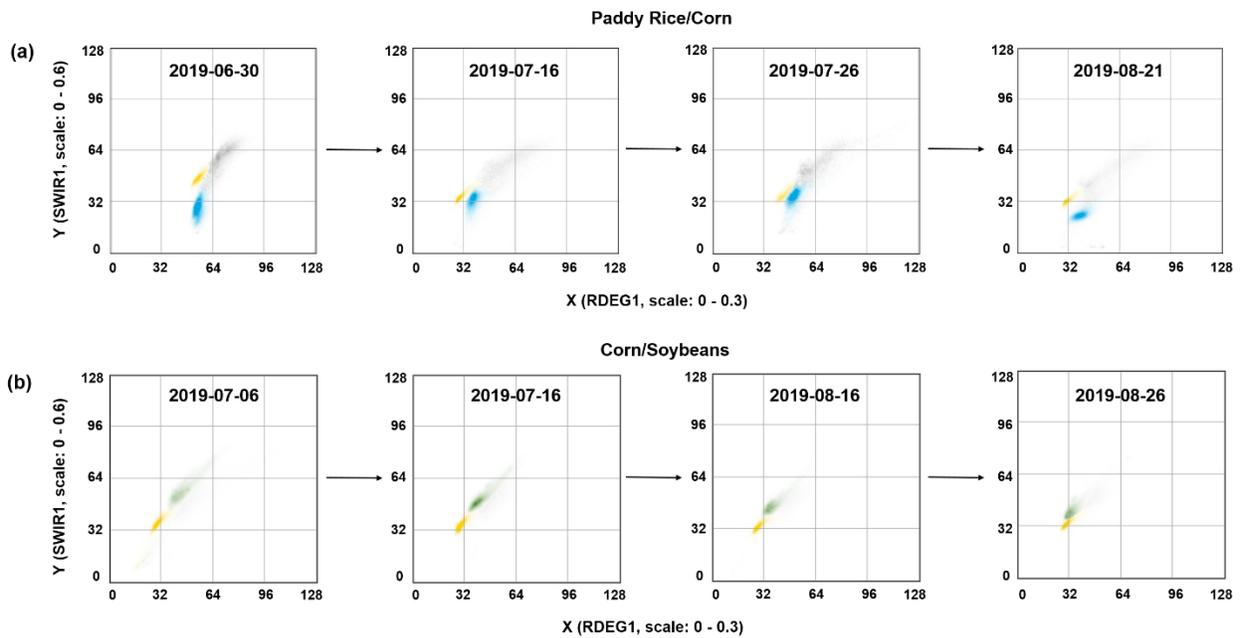

**Fig. 10.** Employed topology relationship between paddy rice (blue)/corn (yellow) and corn (yellow)/soybeans (green). Arrows indicate the progression of the growing season. Note that the date showing the topology relationship can have yearly variation depending on weather conditions, management practice, etc. (a) In the RDEG1/SWIR1 feature space, paddy rice pixels are at the lower right side of corn pixels. (b) In the RDEG1/SWIR1 feature space of Sentinel-2, the corn is first at the bottom left of soybeans and then move directly below soybeans.

**3.2.2. Evaluation of generated labels and classification**

The topology-based approach had good performance in generating sufficient high-quality labels in NE China (Figs. 11a-c). Average agreements of generated labels of paddy rice, corn and soybeans were 0.967 0.985 and 0.915 while the number of labels finally reached 2,265,308, 9,022,148 and 195,993, respectively. Subsequent paddy rice/corn/soybeans classification also achieved high accuracy when evaluated by the field-collected ground truth (Fig. 11d). F1 scores for three crops and the OA exceeded 0.85 since DOY 196 (Jul 15th) and slowly increased until reaching plateaus. The maximum F1 scores for paddy rice, corn and soybeans were 0.892, 0.897, 0.918 and the maximum OA was 0.893. In comparison, the cropland map by You et al. (2021) had a lower F1 score for corn (0.753), F1 score for soybeans (0.834) and OA (0.76), but a slightly higher F1 score for paddy rice (0.941) than the map generated in this study.

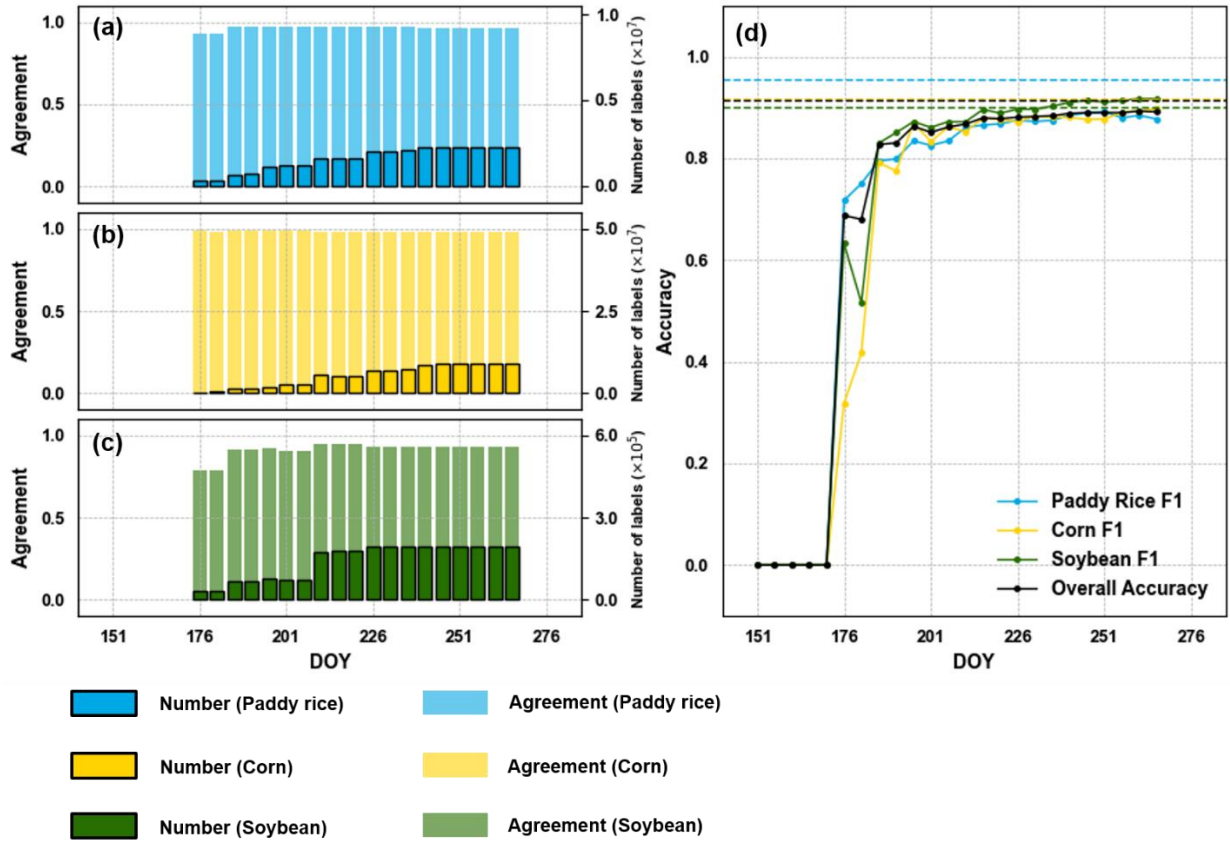

**Fig. 11.** Agreements and number of generated labels and crop mapping accuracy in NE China. Zero values indicate that no labels were generated at the date. (a) Agreement and label number for paddy rice labels in 2019 generated from the RDEG1/SWIR1 feature space of Sentinel-2. (b) Agreement and label number for corn labels in 2019 generated from the RDEG1/SWIR1 feature space of Sentinel-2. (c) Agreement and label number for soybean labels in 2019 generated from the RDEG1/SWIR1 feature space of Sentinel-2. (d) F1 scores and OA of paddy rice/corn/soybeans classification. Four dash lines, i.e., blue, yellow, green and black, indicate F1 scores for paddy rice, corn and soybeans and OA using the cropland map in NE China derived from the postseason mapping.

## 4. Discussion

### 4.1. Interpretation of classification performances: benchmark comparison and further improvement

The proposed topology-based approach differs from traditional decision boundary-based approaches in that it attempts to transfer "topologically stable" features to generate high quality labels in the target year that effectively support subsequent crop type mapping. To further consolidate this statement, we compared our early-season mapping built on labels generated by the topology-based approach to a benchmark approach that used Sentinel-2 imagery from other years to train the RF classifier to build decision boundaries and transferred them to the target year, including (1) 2017&2018 for training and 2019 for testing and (2) 2017&2019 for training and 2018 for testing. Training labels were randomly selected from CDLs in training years. The benchmark approach used the same way of compositing time series images and the same input features as those described in section 2.5. In testing years of both 2018 and 2019, results of this study consistently outperformed the benchmark approach after DOY 176 (Jun 25th) in all three metrics (Fig. 12) as more and more high quality labels were generated. The highest OA was 0.865 by using labels from the proposed approach compared to 0.778 by the benchmark approach in 2018 and 0.907 compared to 0.827 in 2019.

In addition, we compared our results with two postseason classification approaches that used spectral information of the entire growing season in historical years. Similarly, we used (1) 2017&2018 for training and 2019 for testing and (2) 2017&2019 for training and 2018 for testing (3) CDLs in training years to provide labels. The first postseason approach used the full time series of the year to train the RF classifier and transferred it to the target year. This approach achieved an OA of 0.813 in 2018 and 0.873 in 2019. The second approach fitted the harmonic regression to the full-year time series and trained the RF classifier using coefficients of the regression (Wang et al., 2019). The best classification OAs using different hyperparameters were 0.713 in 2018 and 0.853 in 2019, both of which were worse than the accuracy achieved by this study. The results of two postseason approaches revealed that even leveraging the information of the full growing season did not work well without using labels in the target year.

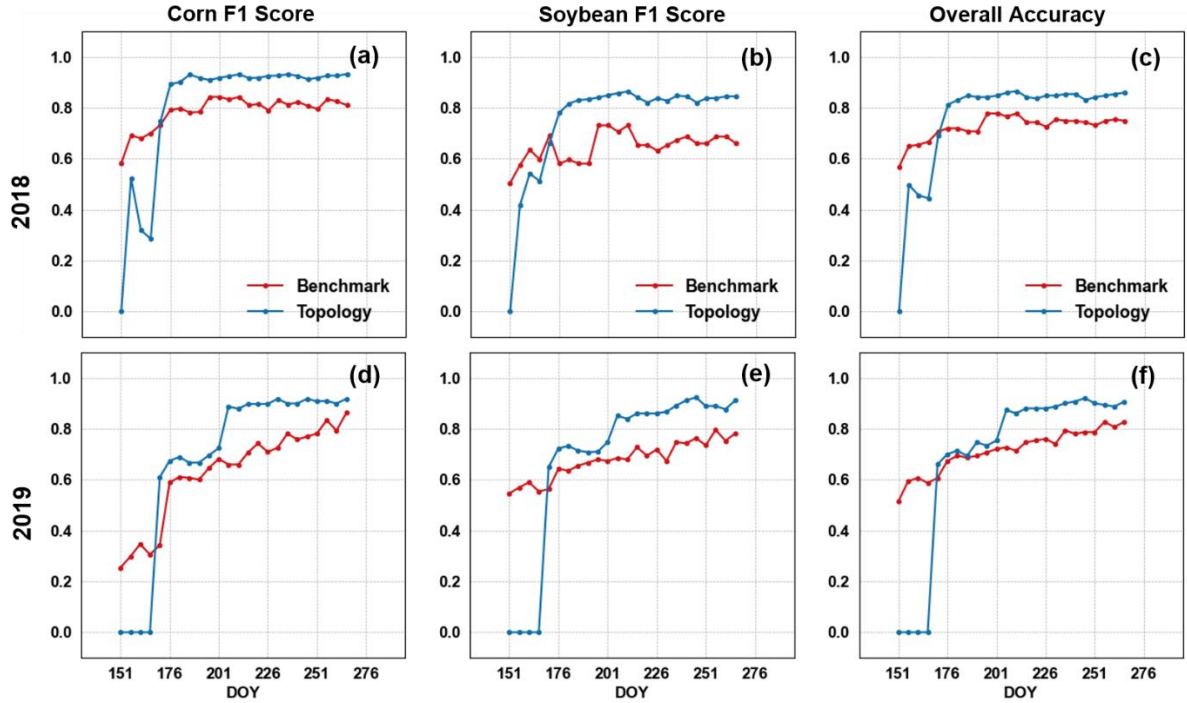

**Fig. 12.** Date-to-date comparison between corn F1 score, soybean F1 score and overall accuracy between the proposed approach using Sentinel-2 NIR/SWIR1 feature space and benchmark approach using Sentinel-2. The benchmark approach transferred historical decision boundaries to the target year to classify corn, soybeans and background. Zero values indicate that no labels were generated at the date. (a) Comparison of F1 score of corn in 2018. (b) Comparison of F1 score of soybean in 2018. (c) Comparison of overall accuracy in 2018. (d) Comparison of F1 score of corn in 2019. (e) Comparison of F1 score of soybean in 2019. (f) Comparison of overall accuracy in 2019.

Though the comparison with the benchmark approach and the best OA (Table 3) already demonstrates the capacity of the topology-based approach in addressing label limitation in crop type mapping, all classifications in this study were based on a simple RF classifier and potential pathways exist to further boost the classification performance. We can improve the performance in at least two ways. First, rather than relying on spectral features from a single-sensor, introducing multi-sensor data, e.g., a combination of two or three from Sentinel-1, Sentinel-2, Landsat-7 and Landsat-8, were proved to be effective to capture more features in crop type mapping tasks (Johnson and Mueller, 2021; Song et al., 2021; Xu et al., 2021, 2020; You and Dong, 2020). Second, replacing the RF classifier with more advanced algorithms, such as artificial neural network (ANN), convolutional neural network (CNN) and long short-term memory network (LSTM), may further improve the classification accuracy. Several recent studies have demonstrated the effectiveness of these algorithms when labels are available (Cai et al., 2018; Xu et al., 2021, 2020; Zhong et al., 2019).

**Table 3**
The best OA achieved during the growing season for different experiments. The last row denotes the mean best OA for all experiments

| Region | Sensor | Feature Space | Year | Best OA |
|---|---|---|---|---|
| Iowa | Sentinel-2 | RDEG1/SWIR1 | 2018 | 0.860 |
| Iowa | Sentinel-2 | RDEG1/SWIR1 | 2019 | 0.873 |
| Iowa | Sentinel-2 | NIR/SWIR1 | 2018 | 0.865 |
| Iowa | Sentinel-2 | NIR/SWIR1 | 2019 | 0.907 |
| Iowa | Landsat-8 | NIR/SWIR1 | 2018 | 0.847 |
| Iowa | Landsat-8 | NIR/SWIR1 | 2019 | 0.895 |
| NE China | Sentinel-2 | RDEG1/SWIR1 | 2019 | 0.893 |
| \ | \ | \ | \ | **0.877** |

## 4.2. Determination of feature spaces and topology relationships

In section 3, we only used two feature spaces (i.e., RDEG1/SWIR1 and NIR/SWIR1) for generating labels, but it does not indicate that the two are the only feasible feature spaces. In fact, other feature spaces may also support the discovery of topological relationships that result in similar prediction performance. Fig. 13 illustrates five possible feature spaces using the Sentinel-2 imagery. The first column of each panel shows the time series of spectral features of corn and soybeans, with yellow and green lines (both solid and dashed) representing mean values of two features for corn and soybeans and the shaded areas indicate one standard deviation. The four orange dashed lines are dates for the subsequent four heat maps.

Take Fig.13a as an example, in DOY 176 (Jun 25th), soybeans have higher values in both RDEG1 and SWIR1 bands than corn. Therefore, along the x-axis (RDEG1), soybeans will locate at the left side of the corn and along the y axis (SWIR1), soybeans are above the corn. In addition, because soybeans and corn both have large variations (represented by the shaded area), they tend to distribute more widely in the heat map and thus have longer extensions along with both the x- and y-axis. In DOY 201 (Jul 20th), while soybeans still have higher values in the two bands, the variations have been narrowed down. Therefore, soybeans and corn keep the same relative position but have more concentrated shapes. In DOY 226 (Aug 14th) and DOY 251 (Sep 8th), soybeans and corn have comparable values in RDEG1 band and similar positions along the x-axis. As soybeans still have larger SWIR1 values, they continue to appear at the upper side of the corn.

The example in Fig. 13 reveals that whether a feature space can present any topology relationship can be straightforwardly observed from the spectral time series, that is: when at least one of the two bands that compose the feature space is distinguishable for corn and soybeans, i.e., the shaded areas have little overlapping, the heat map of that feature space will have a unique topology relationship (Fig. 13a, b, d, e). If corn and soybean spectral time series overlap in both bands, they will also overlap in the heat map and present no topology relationship (Fig. 13c). As a summary, the process of determining feature spaces and topology relationships is similar to determining what spectral features are distinguishable for target classes that are prerequisites in traditional classification tasks.

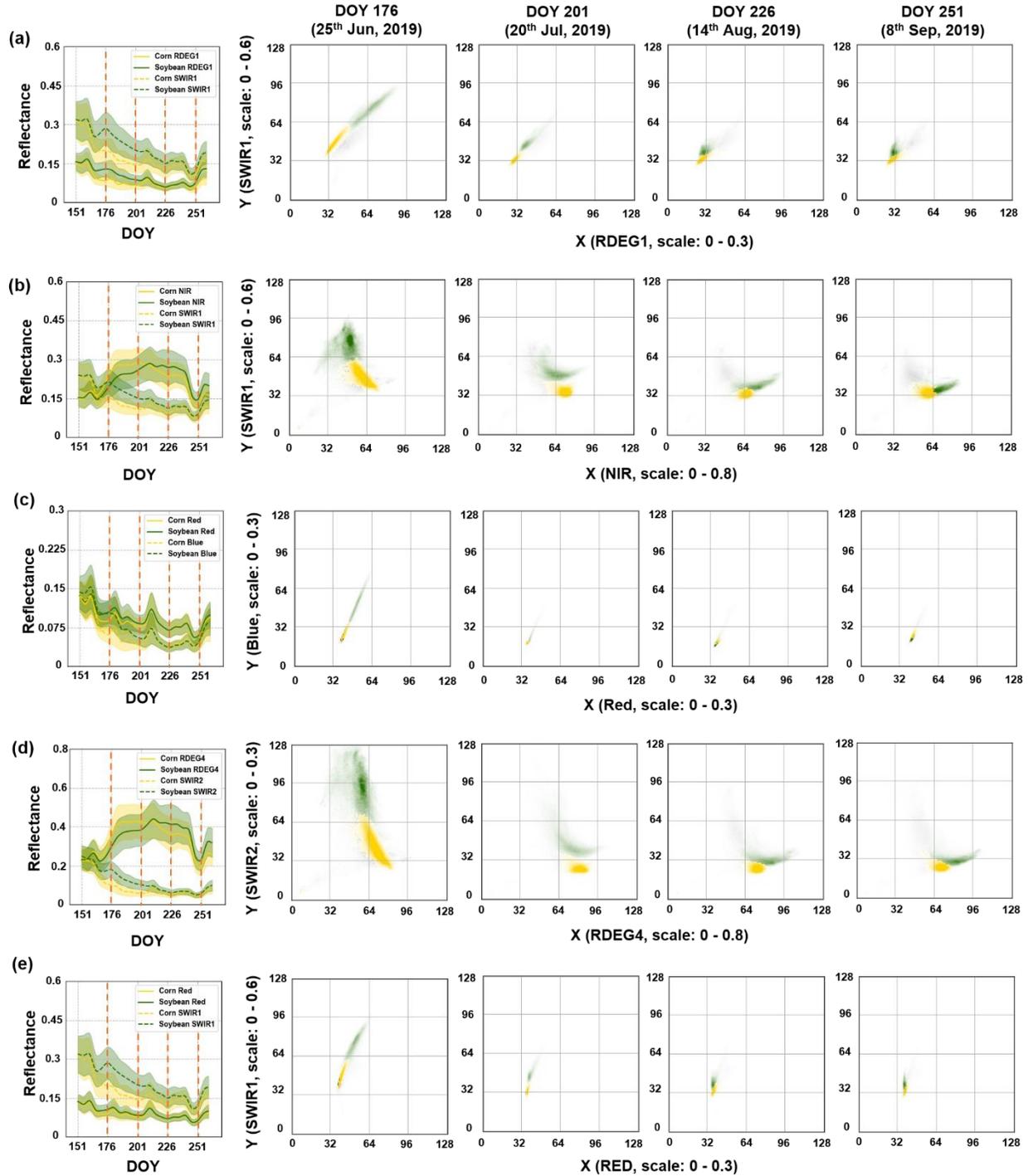

**Fig. 13.** Topology relationships between corn and soybeans in different feature spaces. The first plot in each subfigure shows the time series spectral features of corn and soybeans. The yellow and green lines (both solid and dashed) are mean values of two features for corn and soybeans and the shaded areas indicate one standard deviation. The four orange dashed lines are dates for four heat maps presented in each subfigure. (a) The RDEG1/SWIR1 feature space; (b) The NIR/SWIR1 feature space; (c) The Red/Blue feature space; (d) RDEG14/SWIR12 feature space; (e) The Red/SWIR1 feature space.

### 4.3. The earliest date for crop type mapping

Similar to some previous studies (Cai et al., 2018; Johnson and Mueller, 2021; Konduri et al., 2020; You and Dong, 2020), our results show that the classification performance improved over time. Assuming a threshold exceeding or approximating 0.85 for OA as the acceptable accuracy for the generated maps, the earliest dates for mapping are mostly between early July and early August under different experiments (Table 4). In Iowa, these dates roughly equal to the beginning of the corn silking and soybean flowering stage and are more than two months ahead of harvest according to the USDA NASS crop progress report (CPR) for 2018 and 2019 (Fig. 9). In NE China, the earliest date of DOY 196 (Jul 15th) means that our classification could have an acceptable accuracy at least two and a half months before the harvest of corn and soybean in late September and the harvest of paddy rice in early October (You and Dong, 2020). Our classification framework thus allows a significant time-lead in crop mapping and pre-harvest decision-making. Interestingly, using Sentinel-2 imagery in general had earlier dates than Landsat-8 imagery, which is more due to the frequent temporal resolution of Sentinel-2 instead than the difference in spatial or spectral difference. With Landsat-8, it requires a longer time to accumulate sufficient labels across the study region to better account for the spatial variability. In addition, it also takes longer to acquire sufficient time-series images from Landsat-8 for the classifier to learn robust features and be less vulnerable to outliers such as cloud contamination.

**Table 4**
The earliest date to achieve an OA of around 0.85 for different experiments

| Region | Sensor | Feature Space | Year | Earliest Date |
|---|---|---|---|---|
| Iowa | Sentinel-2 | RDEG1/SWIR1 | 2018 | DOY 191 (Jul 10th) |
| Iowa | Sentinel-2 | RDEG1/SWIR1 | 2019 | DOY 206 (Jul 25th) |
| Iowa | Sentinel-2 | NIR/SWIR1 | 2018 | DOY 186 (Jul 5th) |
| Iowa | Sentinel-2 | NIR/SWIR1 | 2019 | DOY 206 (Jul 25th) |
| Iowa | Landsat-8 | NIR/SWIR1 | 2018 | DOY 231 (Aug 19th) |
| Iowa | Landsat-8 | NIR/SWIR1 | 2019 | DOY 215 (Aug 3rd) |
| NE China | Sentinel-2 | RDEG1/SWIR1 | 2019 | DOY 196 (Jul 15th) |

### 4.4. Limitations and future plans

The proposed method showed the capacity of transferring historical knowledge and generating within-year labels to support further classification in the target year. However, some limitations remain to be addressed in future studies. First, in the study, we only aimed to recognize topological relationships between two crops, e.g., corn/soybeans and paddy rice/corn. How applicable our approach is to multiple crop types (e.g. 5+) requires further exploration. Second, topological relationships are somewhat sensitive to the relative abundance of target classes. We noted that when target crops have a very small portion in the 512×512 image patch, the background is likely to obscure them in the heat map. This is less of a concern for major crops since we can always find sufficient image patches where they account for the majority of the region. However, for some minor crops, e.g., spring wheat, barley and oats etc., that have small cultivated areas but are of equal interest, the background may overshadow the target class. Third, CDL in Iowa and cropland map in NE China provided pixel-wise pseudo ground truth in training years and played a major role in heat map generation. In regions where similar products are not available, e.g., many underdeveloped countries in Africa, the implementation of the proposed approach can be largely limited.

Our future plan will focus on addressing the three limitations mentioned above. First, to extend to multiclass classification, stacking multiple 2D feature spaces or trying 3D feature spaces can be potentially helpful. Feature spaces with dimensionality higher than 3D may not be considered at this moment because of the difficulty to visualize it. Second, to address its dependence on the percentage of target classes, new ways of constructing heat maps can be used. For example, one potential way would be drawing points randomly over the entire study area to compose the heat map rather than splitting the study area into image patches and generating heat maps based on those patches. Third, to apply the approach to regions without well-recognized products such as CDL, self-supervised learning and few-shot learning can play a role in creating simple land cover maps first. Even though the land cover map generated through self-supervised learning and few-shot learning might have unsatisfactory accuracy, it provides reference data and can help the approach move forward in those regions to some extent (Ghosh et al., 2021; Rußwurm et al., 2020).

## 5. Conclusions and outlook

In this study, we proposed a novel classification framework that mimics object detection tasks in computer vision based on the topological relationships among targets in the feature space to transfer historical knowledge to the target year. Experiments were conducted to test the applicability of the proposed approach with different crop types, sensors, years and regions. Results indicated that the proposed method was able to generate high-quality labels under different scenarios. In addition, subsequent crop type mapping with a simple RF classifier based on generated labels revealed that the OA for corn/soybeans mapping in Iowa could reach 0.8 as early as the beginning of the silking or flowering stage in the target year, bringing the available cropland map at least two months earlier than the harvest. Similar results could also be found in paddy rice/corn/soybeans mapping in NE China. Our next step will extend it to other regions with even more challenging situations in ground truth available by incorporating other advanced deep learning algorithms such as self-supervised learning and few-shot learning.

**Supplementary Material**

Please refer to:

https://docs.google.com/document/d/1-GVSvzS7xY2wJpH3P-wv0oLSDH_BLOgF/edit?usp=sharing&ouid=108579315095106831824&rtpof=true&sd=true